\documentclass[11pt]{article}

\usepackage[margin=1in]{geometry}
\usepackage{amsmath,amssymb}
\usepackage{url}
\usepackage{xcolor}
\usepackage{graphicx}
\usepackage{booktabs}
\usepackage{natbib}

\title{Predicting Mortgage Default with Machine Learning: AutoML, Class Imbalance, and Leakage Control}

\author{
  Xianghong Hu\thanks{Amazon Web Services, Inc. \ \ \texttt{hxianglo@amazon.com}}
  \and
  Tianning Xu\thanks{University of Illinois Urbana-Champaign. \ \ \texttt{tx8@illinois.edu}}
  \and
  Ying Chen\thanks{University of Connecticut. \ \ \texttt{yingchen.cheryl@gmail.com}}
  \and
  Shuai Wang\thanks{University at Buffalo, SUNY. \ \ \texttt{swang53@buffalo.edu}}
}

\date{}

\begin{document}
\maketitle

\begin{abstract}
Mortgage default prediction is a core task in financial risk management, and machine learning models are increasingly used to estimate default probabilities and provide interpretable signals for downstream decisions. In real-world mortgage datasets, however, three factors frequently undermine evaluation validity and deployment reliability: ambiguity in default labeling, severe class imbalance, and information leakage arising from temporal structure and post-event variables. We compare multiple machine learning approaches for mortgage default prediction using a real-world loan-level dataset, with emphasis on leakage control and imbalance handling. We employ leakage-aware feature selection, a strict temporal split that constrains both origination and reporting periods, and controlled downsampling of the majority class. Across multiple positive-to-negative ratios, performance remains stable, and an AutoML approach (AutoGluon) achieves the strongest AUROC among the models evaluated. An extended and pedagogical version of this work will appear as a book chapter.
\end{abstract}


\section{Introduction}

Machine learning methods are increasingly used in mortgage origination and underwriting, complementing or replacing earlier default and risk models. Given the scale of mortgage lending and the downstream roles of underwriting, risk transfer, and post-origination monitoring, it is important to understand which modeling and evaluation choices materially affect the reliability of mortgage default prediction.

This study isolates two factors that strongly influence reported performance in practice. \emph{First}, default prediction is highly imbalanced: non-defaults vastly outnumber defaults. Training on raw data can yield models biased toward the majority class \citep{kim2022empirical}. We use downsampling of the majority class and vary the non-default-to-default ratio (from 100:1 down to 5:1 or 1:1). Results indicate that predictive performance remains stable across these regimes. \emph{Second}, information leakage can inflate validation metrics while degrading generalization. The Fannie Mae Single-Family Loan Performance Data\footnote{\url{https://capitalmarkets.fanniemae.com/credit-risk-transfer/single-family-credit-risk-transfer/fannie-mae-single-family-loan-performance-data}} is especially prone to leakage: several fields are populated only after delinquency or are strongly correlated with post-origination outcomes. We mitigate leakage by (i) dropping high-risk variables—including most IDs, dates, and payment-history fields—and (ii) applying a strict temporal split defined on both origination and reporting (action) dates, so that training, validation, and test sets do not overlap in time.

\subsection{Related Work}

Post-2008, econometric models dominated default prediction, emphasizing interpretable predictors and causal structure. More recently, machine learning has been used to improve predictive performance and support operational decisions \citep{huo2025enhancing}. Common choices include logistic regression, decision trees, and random forests \citep{james2013introduction, breiman2001random, 2018wager:InfJack, xu2024variance}; gradient boosting methods such as XGBoost \citep{chen2015xgboost, chen2016xgboost} and LightGBM \citep{ke2017lightgbm, wang2022corporate}; and AutoML systems \citep{yao2018taking, he2021automl, thornton2013auto} like AutoGluon \citep{erickson2020autogluon}, which automate preprocessing, model selection, and ensembling. We build on this line of work by systematically comparing these approaches on Fannie Mae data under explicit leakage control and imbalance handling.

\section{Methodology}

We use the Fannie Mae Single-Family Loan Performance dataset, which spans over 30 years and includes 110 fields. We restrict to a curated subset of variables that are both informative and leakage-safe. Below we summarize the dataset, preprocessing, partitioning, and models.

\subsection{Dataset and Labeling}

We use Fannie Mae Single-Family Loan Performance data from 2023 Q1 through 2024 Q4. Fannie Mae (Federal National Mortgage Association) is a U.S. government-sponsored entity in the secondary mortgage market and publishes quarterly acquisition and performance updates. Each loan has a unique acquisition window; the same loan can appear in multiple rows because performance is reported monthly. Rows are uniquely identified by \texttt{LOAN\_ID}, \texttt{ORIG\_DATE}, and \texttt{ACT\_PERIOD} (format MMYYYY). We label rows with \texttt{DLQ\_STATUS = 0} as non-default (negative) and \texttt{DLQ\_STATUS > 0} as default (positive).

\subsection{Temporal Partitioning and Leakage Mitigation}
\label{subsec:data_process}

\begin{figure}[htbp]
    \centering
    \includegraphics[width=0.6\linewidth]{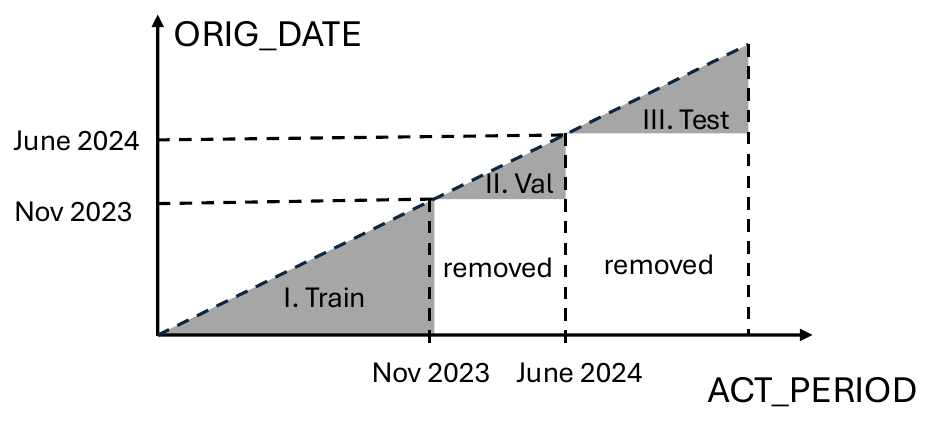}
    \caption{Temporal partition by \texttt{ORIG\_DATE} and \texttt{ACT\_PERIOD}.}
    \label{fig:split}
\end{figure}

Splitting only by \texttt{ORIG\_DATE} or only by \texttt{ACT\_PERIOD} can induce leakage, since both dimensions carry time information. We therefore impose cutoffs on \emph{both} \texttt{ORIG\_DATE} and \texttt{ACT\_PERIOD} and drop any sample that does not fall unambiguously into one of three subsets. By definition, \texttt{ORIG\_DATE} is always earlier than or equal to \texttt{ACT\_PERIOD}; the valid region thus corresponds to a lower-triangular structure in the (origination, action-period) plane. We discard the two rectangular “overlap” regions (see Figure~\ref{fig:split}) and keep three disjoint triangles:

\begin{itemize}
    \item \textbf{Training}: \texttt{ORIG\_DATE} and \texttt{ACT\_PERIOD} both before November 2023.
    \item \textbf{Validation}: both in [November 2023, June 2024).
    \item \textbf{Testing}: both in June 2024 or later.
\end{itemize}

\subsection{Feature Selection and Transformation}

We retain 26 numeric and 16 categorical features. We drop identifiers, timestamps, and payment-history variables to avoid leakage, and exclude categoricals with very high missing rates or cardinality (e.g., $>500$ distinct values). For categoricals we encode the first three digits of the ZIP code via latitude/longitude. LightGBM and AutoGluon receive raw categoricals; other models use one-hot encoding.

\subsection{Downsampling}

The data are strongly imbalanced (roughly 100 non-defaults per default). We fix 19{,}807 positive (default) samples for training, 9{,}932 for validation, and 8{,}987 for testing, then downsample negatives by randomly drawing $x \times$ (number of positives) for $x \in \{1,2,5,10\}$ in each split.

\subsection{Models}

We compare Logistic Regression (L1 and L2), Random Forest, XGBoost, LightGBM \citep{Lin2024Research}, and AutoGluon. Logistic regression models default probability via a logistic link and maximum-likelihood estimation. Random Forest builds an ensemble of trees on bootstrapped samples and random feature subsets. XGBoost and LightGBM use gradient boosting with shallow trees and regularization. AutoGluon \citep{erickson2020autogluon} is an open-source AutoML framework for tabular, multimodal, and time-series tasks \citep{he2021automl}; it automates preprocessing, model selection, tuning, and ensembling. We use its \texttt{TabularPredictor} for our classification task.

\section{Experimental Results}

We train all models on the temporally partitioned data (Section~\ref{subsec:data_process}), tune AutoGluon on the validation set, and use validation-based early stopping for XGBoost and LightGBM. We downsample negatives in training and validation to positive-to-negative ratios 1:1, 1:2, 1:5, and 1:10, and report test-set Area Under the ROC Curve (AUROC) \citep{james2013introduction}, which is less sensitive to class imbalance than precision or recall.

Table~\ref{tab:model_performance} reports test AUROC for each model and ratio. Figure~\ref{fig:roc} plots ROC curves at the 1:2 ratio. Training and validation AUROCs are in Appendix~\ref{append:val_test_performance}.

\begin{table}[htbp]
\centering
\begin{tabular}{lcccc}
\toprule
\textbf{Model \textbackslash Ratio} & \textbf{1:1} & \textbf{1:2} & \textbf{1:5} & \textbf{1:10} \\
\midrule
Logistic Regression L1 & 0.7169 & 0.7240 & 0.7236 & 0.7224 \\
Logistic Regression L2 & 0.7157 & 0.7229 & 0.7231 & 0.7220 \\
XGBoost                & 0.8122 & 0.8112 & 0.8137 & 0.8132 \\
Random Forest          & 0.7962 & 0.7960 & 0.7916 & 0.7864 \\
LightGBM               & 0.8086 & 0.8107 & 0.8147 & 0.8127 \\
AutoGluon              & \textbf{0.8204} & \textbf{0.8230} & \textbf{0.8203} & \textbf{0.8201} \\
\bottomrule
\end{tabular}
\caption{Test AUROC under different positive-to-negative downsampling ratios.}
\label{tab:model_performance}
\end{table}

\begin{figure}[htbp]
    \centering
    \includegraphics[width=0.6\linewidth]{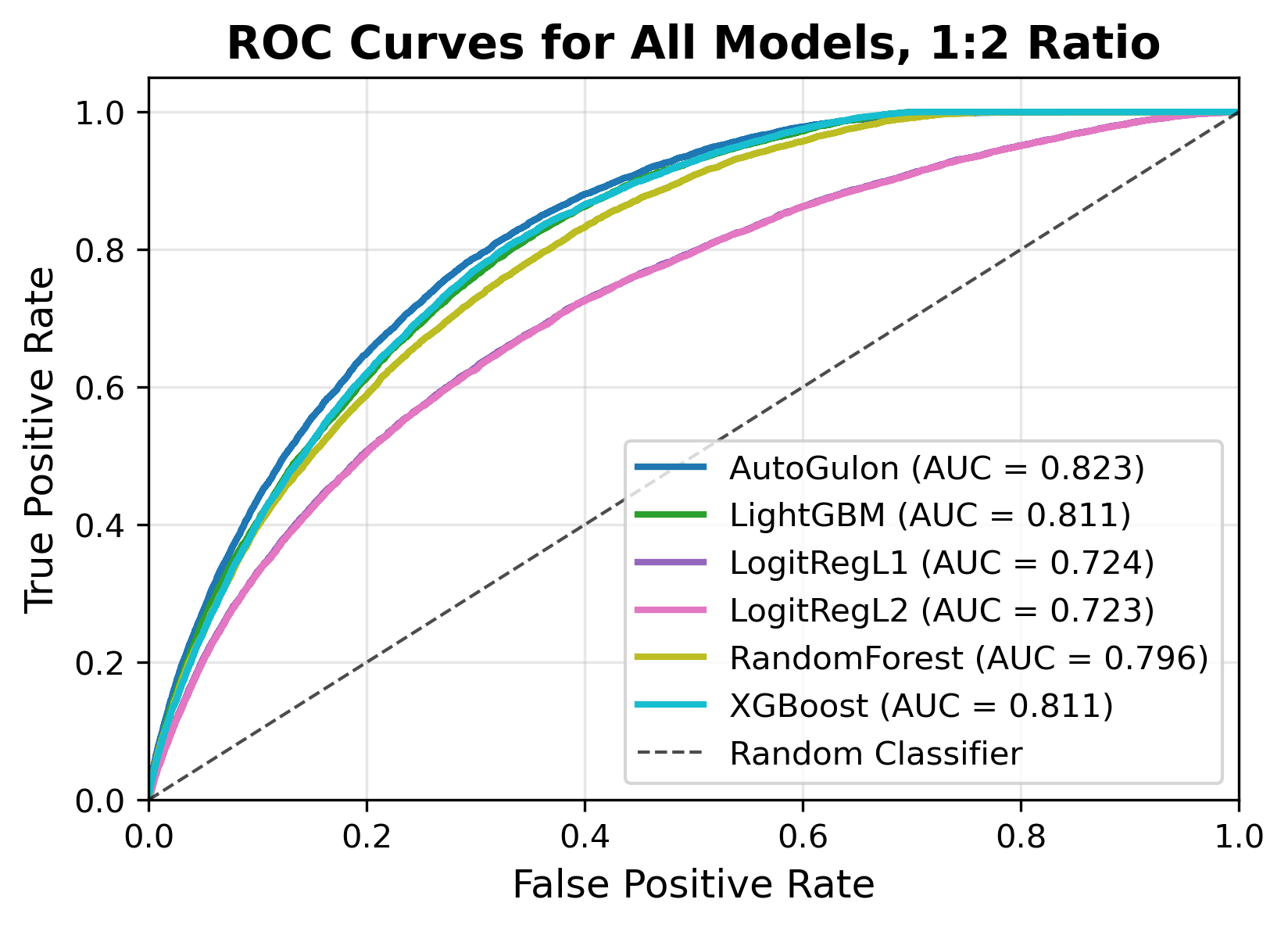}
    \caption{ROC curves at 1:2 positive-to-negative ratio.}
    \label{fig:roc}
\end{figure}

AutoGluon achieves the highest AUROC at every ratio, with XGBoost and LightGBM about 1--2 points lower. The gap likely reflects AutoGluon’s built-in tuning and ensembling; our XGBoost and LightGBM use only early stopping, so further gains from Bayesian optimization or other AutoML tools are possible \citep{wu2019hyperparameter}.

Across downsampling levels, performance is relatively stable. With a fixed number of positives, adding more negatives does not uniformly improve AUROC. AutoGluon peaks at 1:2; XGBoost at 1:5.

At 1:2, AutoGluon reaches training AUROC 0.9567 and test AUROC 0.8230; XGBoost reaches 0.8902 and 0.8112 (see Appendix~\ref{append:val_test_performance}). The larger train--test gap for AutoGluon suggests greater overfitting, consistent with its flexibility. The strict temporal split (Section~\ref{subsec:data_process}) is important to limit leakage and overfitting.

We inspect AutoGluon’s feature importance at 1:2. The four variables with $>5\%$ importance are: (1) months since origination, (2) primary borrower credit score, (3) current unpaid principal balance, and (4) original unpaid principal balance. Importance can vary by model and correlation structure; the top-30 list is in Appendix~\ref{append:val_test_performance}.

\section{Conclusion}

We compare Logistic Regression, Random Forest, XGBoost, LightGBM, and AutoGluon on Fannie Mae single-family loan data for mortgage default prediction. AutoGluon attains the best AUROC across all positive-to-negative ratios we consider. We downsample the non-default class to ease imbalance and reduce compute; results show that downsampling preserves or slightly improves performance while cutting cost. Controlling leakage via careful feature selection and a strict temporal split is important for reliable evaluation.

\bibliographystyle{plainnat}
\bibliography{reference}

\appendix
\section{Additional Experimental Results}
\label{append:val_test_performance}

Tables~\ref{tab:train_auroc} and~\ref{tab:val_auroc} report training and validation AUROC across positive-to-negative ratios.

\begin{table}[htbp]
\centering
\begin{tabular}{lcccc}
\toprule
\textbf{Model \textbackslash Ratio} & \textbf{1:1} & \textbf{1:2} & \textbf{1:5} & \textbf{1:10} \\
\midrule
Logistic Regression L1 & 0.7663 & 0.7674 & 0.7665 & 0.7660 \\
Logistic Regression L2 & 0.7667 & 0.7676 & 0.7665 & 0.7661 \\
XGBoost                & 0.8902 & 0.8847 & 0.8766 & 0.8665 \\
Random Forest          & 0.8927 & 0.8867 & 0.8865 & 0.8905 \\
LightGBM               & 0.8932 & 0.8797 & 0.8758 & 0.8639 \\
AutoGluon              & 0.9572 & 0.9567 & 0.9335 & 0.9299 \\
\bottomrule
\end{tabular}
\caption{Training AUROC under different downsampling ratios.}
\label{tab:train_auroc}
\end{table}

\begin{table}[htbp]
\centering
\begin{tabular}{lcccc}
\toprule
\textbf{Model \textbackslash Ratio} & \textbf{1:1} & \textbf{1:2} & \textbf{1:5} & \textbf{1:10} \\
\midrule
Logistic Regression L1 & 0.7167 & 0.7235 & 0.7231 & 0.7222 \\
Logistic Regression L2 & 0.7152 & 0.7223 & 0.7225 & 0.7216 \\
XGBoost                & 0.8147 & 0.8182 & 0.8143 & 0.8149 \\
Random Forest          & 0.7925 & 0.7965 & 0.7898 & 0.7863 \\
LightGBM               & 0.8114 & 0.8191 & 0.8153 & 0.8159 \\
AutoGluon              & 0.8257 & 0.8299 & 0.8251 & 0.8233 \\
\bottomrule
\end{tabular}
\caption{Validation AUROC under different downsampling ratios.}
\label{tab:val_auroc}
\end{table}

\begin{figure}[htbp]
    \centering
    \includegraphics[width=1\linewidth]{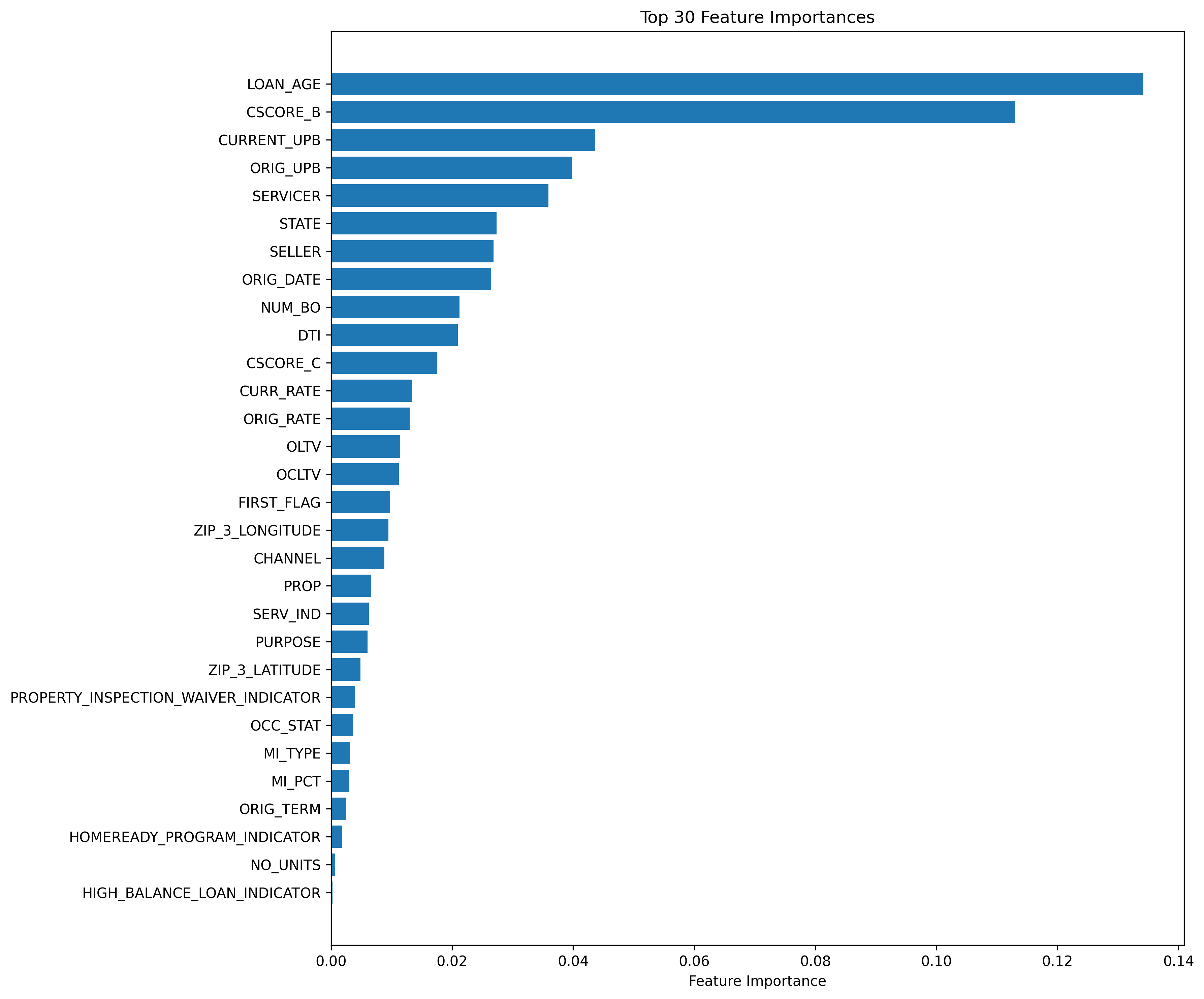}
    \caption{AutoGluon feature importance at 1:2 ratio.}
    \label{fig:fi_ag}
\end{figure}

\section{Model Hyperparameters}

Logistic Regression and Random Forest use \texttt{scikit-learn}; XGBoost and LightGBM use their eponymous Python packages; AutoGluon uses the tabular module. All runs use the same A100 GPU environment.

Logistic Regression: L1 and L2 with \texttt{C = 1}. Random Forest: 150 trees, \texttt{min\_samples\_split = 8}, \texttt{max\_depth = 15}, Gini split criterion. XGBoost and LightGBM: \texttt{num\_iterations = 200}, \texttt{learning\_rate = 0.05}, \texttt{max\_depth = 8}, validation-based early stopping. AutoGluon: \texttt{presets = medium} (covers architecture, ensembling, and tuning).

\end{document}